\documentclass{article}

\usepackage[main, final]{neurips_2025}
 \usepackage{amsmath}
 \usepackage{graphicx}
 \usepackage{subcaption}

\usepackage[utf8]{inputenc} % allow utf-8 input
\usepackage[T1]{fontenc}    % use 8-bit T1 fonts
\usepackage{hyperref}       % hyperlinks
\usepackage{url}            % simple URL typesetting
\usepackage{booktabs}       % professional-quality tables
\usepackage{amsfonts}       % blackboard math symbols
\usepackage{nicefrac}       % compact symbols for 1/2, etc.
\usepackage{microtype}      % microtypography
\usepackage{xcolor}         % colors

\title{Neural Triangular Transport Maps: A New Approach Towards Sampling in Lattice QCD}

\author{%
  Andrey Bryutkin \\
  Department of Mathematics\\
  Massachusetts Institute of Technology\\
  \texttt{bryutkin@mit.edu} \\
  \And
  Youssef Marzouk \\
  Laboratory for Information and Decision Systems \\
  Massachusetts Institute of Technology \\
  \texttt{ymarz@mit.edu} \\
}
\begin{document}
\maketitle
\begin{abstract}
Lattice field theories are fundamental testbeds for computational physics; yet, sampling their Boltzmann distributions remains challenging due to multimodality and long-range correlations. While normalizing flows offer a promising alternative, their application to large lattices is often constrained by prohibitive memory requirements and the challenge of maintaining sufficient model expressivity. We propose sparse triangular transport maps that explicitly exploit the conditional independence structure of the lattice graph under periodic boundary conditions using monotone rectified neural networks (MRNN). We introduce a comprehensive framework for triangular transport maps that navigates the fundamental trade-off between \emph{exact sparsity} (respecting marginal conditional independence in the target distribution) and \emph{approximate sparsity} (computational tractability without fill-ins). Restricting each triangular map component to a local past enables site-wise parallel evaluation and linear time complexity in lattice size $N$, while preserving the expressive, invertible structure. Using $\phi^4$ in two dimensions as a controlled setting, we analyze how node labelings (orderings) affect the sparsity and performance of triangular maps. We compare against Hybrid Monte Carlo (HMC) and established flow approaches (RealNVP).
\end{abstract}

\section{Introduction}
Lattice field theories provide a non-perturbative framework for fundamental physics, but their study is often constrained by the computational cost of sampling from the high-dimensional Boltzmann distribution, $P[\phi] \propto e^{-S[\phi]}$. While standard MCMC methods like Hybrid Monte Carlo (HMC) are asymptotically exact, they are significantly affected by critical slowing down near phase transitions, where autocorrelation times grow rapidly (often polynomially with the correlation length) \citep{sokal1997monte}.

To address this bottleneck, normalizing flows (NFs) have recently been introduced to lattice field theory, demonstrating the potential for improvements in sampling efficiency \citep{albergo2019flow, albergo2021intro, boyda2021sungauge}. However, existing architectures struggle to scale to large volumes because capturing long-range correlations typically requires many sequential coupling/transform layers, making inference depth-limited \citep{abbott2022aspectsscalingscalabilityflowbased}. In this work, we propose a solution by developing triangular transport maps, which leverage the use of the Knothe-Rosenblatt rearrangements.

The sparse version of our method achieves linear scaling in lattice size $N$. We accomplish this by constraining each output of the map to depend only on a local neighborhood of preceding variables, determined by a specific node ordering. This approach navigates the crucial trade-off between exact sparsity, which requires modeling computationally expensive "fill-in" effects from marginalization, and approximate sparsity, which enforces strict locality for tractability. Any bias from approximate sparsity is removed by a final Metropolis–Hastings correction, yielding asymptotically exact samples under standard MCMC regularity conditions. Each map component is parameterized as a monotone rectified neural network (MRNN), which is an integral of a strictly positive neural network that ensures invertibility and a tractable Jacobian.

\paragraph{Contributions:} Our key contributions are as follows:
\begin{enumerate}
    \item We use Monotone Rectified Neural Networks (MRNNs) to flexibly define the map's invertible components. This avoids the rigidity of polynomial-based methods, which require setting the polynomial degree in advance.
    \item We introduce a conditionally sparse triangular map—a transformation where dependencies are restricted to local neighborhoods — for lattice models using MRNN components, reducing the computational and memory complexity to $\mathcal{O}(N)$.
    \item We empirically evaluate several labeling strategies and their impact on sparsity patterns and sampling efficiency.
    \item We empirically demonstrate that, in the 2D $\phi^4$ theory, the method shows competitive sampling efficiency, establishing a framework for future extensions to gauge theories.
\end{enumerate}

\section{Related Work}
Pioneering work by \citep{albergo2019flow} introduced normalizing flows to lattice physics using the RealNVP architecture \citep{dinh2017realnvp}, later refining the model by replacing the initial dense MLP coupling layers with convolutional neural networks to enforce parameter sharing and improve scalability \citep{albergo2021intro}. These models are computationally efficient but must be stacked deeply to capture long-range physical correlations. A parallel line of work enforces physical symmetries via gauge-equivariant flows \citep{kanwar2020equivariant, Boyda_2021}. Continuous normalizing flows have also been explored, but they require ODE solvers that add cost \citep{bauer2024superresolvingnormalisingflowslattice}. Despite this progress, developing flow architectures that are both expressive and scalable to large, physically relevant lattices remains a central challenge for the field \citep{abbott2025progressnormalizingflows4d}.

An alternative to coupling-based architectures are triangular transport maps \citep{marzouk2016sampling, Baptista_2023, ramgraber2025friendlyintroductiontriangulartransport}, which correspond to autoregressive models. These models are known to be universal approximators and are used for density estimation \citep{huang2018neural, wang2022minimaxdensityestimationmeasure}. However, their power comes with a critical drawback for sampling: generating a single sample is a sequential process with $\mathcal{O}(N^2)$ complexity, making dense autoregressive models slow for large systems. Our work targets this sampling bottleneck. Recent work has explored monotone parameterizations using polynomials and structure exploitation \citep{Baptista_2023, brennan2020greedy}. In our case, we want to construct maps that respect the conditional independence structure of the target distribution, using the relation between graphical models and conditional independence \citep{spantini2018inferencelowdimensionalcouplings}. The primary obstacle is the phenomenon of "fill-in," where marginalizing variables introduce dense dependencies not present in the original model. This is a classic challenge known as the minimum fill-in problem in the context of sparse Cholesky factorization in numerical linear algebra \citep{george1981computer} and as a core problem for exact inference in probabilistic graphical models \citep{koller2009probabilistic}.

\section{The $\phi^4$ Lattice Field Theory}
Consider a $D$-dimensional hypercubic lattice $\Lambda=(\mathbb{Z} / L \mathbb{Z})^D$ with $N = L^D$ sites, where $L$ is the extent in each dimension. A scalar field $\phi_x\in\mathbb{R}$ is defined at each site $x \in \Lambda$. The Euclidean action for the $\phi^4$ theory is given by:
\begin{equation}
S[\phi] = \sum_{x \in \Lambda} \left[ \frac{1}{2} \sum_{\mu=1}^{D} (\phi_{x+\hat{\mu}} - \phi_x)^2 + \frac{m_0^2}{2} \phi_x^2 + \frac{\lambda_0}{4!} \phi_x^4 \right]
\label{eq:action}
\end{equation}
where $\phi_{x+\hat{\mu}}$ is the field at the site adjacent to $x$ in the positive $\mu$-direction, $m_0^2$ is the bare mass-squared parameter, and $\lambda_0$ is the bare coupling constant. We assume periodic boundary conditions $\phi_{x+L\hat{\mu}} = \phi_x$. The induced Gibbs distribution is a strictly positive density forming a Markov random field (MRF) with cliques given by on-site and nearest-neighbor interactions:
\begin{equation}
P[\phi] = \frac{1}{Z} e^{-S[\phi]}
\label{eq:boltzmann}
\end{equation}
where $Z = \int \mathcal{D}\phi \, e^{-S[\phi]}$ is the partition function and $\mathcal{D}\phi = \prod_{x \in \Lambda} d\phi_x$. This Markov property, $\phi_x \perp \phi_{\Lambda \setminus (\{x\} \cup \mathcal{N}(x))} | \phi_{\mathcal{N}(x)}$, is key to our sparse construction.

\section{Triangular Maps}
Transport maps learn a diffeomorphism $T: \mathcal{Z} \to \mathcal{X}$ between a reference distribution $p_Z(z)$ (e.g., a standard $D \cdot N$-dimensional Gaussian) over $z \in \mathcal{Z}$ and a target distribution $p_{\Phi}(\phi)$ (approximating $P[\phi]$) over $\phi \in \mathcal{X}$. The KR-type maps are uniquely defined once an ordering of coordinates is chosen. In general, orderings impact both the expressivity and the computational structure, especially if we look closely into approximate sparsity. If $\phi = T(z)$, the change of variables formula gives:
\begin{equation}
p_{\Phi}(\phi) = p_Z(T^{-1}(\phi)) \left| \det J_{T^{-1}}(\phi) \right|
\end{equation}
or, equivalently, for $z = T^{-1}(\phi)$, $p_{\Phi}(T(z)) = p_Z(z) \left| \det J_T(z) \right|^{-1}$, where $J_T(z)$ is the Jacobian matrix of the transformation $T$ at $z$. We impose an ordering on the $N$ components of $z = (z_0, \dots, z_{N-1})$ and $\phi = (\phi_0, \dots, \phi_{N-1})$. A map is triangular map if each output component $\phi_j$ depends only on its corresponding input $z_j$ and all preceding input components $z_{<j}$. Formally, the map $T: \mathcal{Z} \rightarrow \mathcal{X}$ is defined component-wise as $\phi=T(z)$, where each component $\phi_j$ is generated as:
\begin{align}
\phi_0 &= T_0(z_0) \nonumber \\
\phi_1 &= T_1(z_1; z_0) \nonumber \\
&\vdots \nonumber \\
\phi_j &= T_j(z_j; z_0, z_1, \dots, z_{j-1}) \quad \text{or simply } T_j(z_j; z_{<j}) \nonumber \\
\label{eq:triangular_map_components}
\end{align}
In this structure, each component function $T_j$ applies a transformation to its primary input, $z_j$, where a semicolon separates the primary variable from the conditioning variables. This transformation is specifically designed to be monotonic and invertible with respect to $z_j$. All preceding variables, $z_{<j}=\left(z_0, \ldots, z_{j-1}\right)$, act as a conditioning context, essentially setting the parameters for how $z_j$ is transformed. This autoregressive definition ensures that the Jacobian matrix of the transformation, $J_T(z)$ with elements $(J_T)_{ij} = \frac{\partial \phi_i}{\partial z_j}$, is lower triangular. The determinant is then simply the product of the diagonal entries: $\det J_T(z) = \prod_{j=0}^{N-1} \frac{\partial \phi_j}{\partial z_j}$. A specific parameterization for each component $T_j$ that ensures invertibility with respect to $z_j$ and a positive partial derivative $\frac{\partial \phi_j}{\partial z_j}$ is the monotone rectified component:
\begin{equation}
\phi_j = T_j(z_j ; z_{<j}) = f_j(z_{<j}) + \int_0^{z_j} r(g_j(s, z_{<j})) ds
\label{eq:component_form_integral}
\end{equation}
Here, $r: \mathbb{R} \rightarrow \mathbb{R}^{+}$is a rectifier function that enforces a positive integrand (e.g., $r(s) = \exp(s)$ or Softplus). The core of our method is to parameterize the shift function $f_j$ and the scale integrand $g_j$ using neural networks. This parameterization using monotone rectified neural networks (MRNN) differs from earlier approaches based on orthogonal polynomial expansions \citep{Baptista_2023}. Neural parameterizations are adaptive; by universal approximation results, they can match the expressive guarantees of polynomial maps while avoiding explicit degree selection.

The above parametrization ensures $\frac{\partial \phi_j}{\partial z_j} > 0$. The partial derivative required for the Jacobian determinant is $\frac{\partial \phi_j}{\partial z_j} = r(g_j(z_j, z_{<j}))$. Thus, the log-determinant of the full map $T$ is $\log |\det J_T(z)| = \sum_{j=0}^{N-1} \log r(g_j(z_j, z_{<j}))$. The integral in Eq. \eqref{eq:component_form_integral} is one-dimensional and can be approximated using a change of variables and numerical quadrature:
\begin{equation}
\int_0^{z_j} r(g_j(s, z_{<j})) ds = z_j \int_0^1 r(g_j(t z_j, z_{<j})) dt \approx z_j \sum_{q=1}^{Q} w^{(q)} r(g_j(t^{(q)} z_j, z_{<j}))
\end{equation}
where $(w^{(q)}, t^{(q)})$ are the weights and nodes of a chosen quadrature rule (e.g., Gauss-Legendre) on $[0,1]$, and $Q$ is the number of quadrature points. This formulation is highly efficient for GPU computation. For a batch of size $B$, the evaluation of $g_j$ can be parallelized across the $B \times Q$ inputs corresponding to the quadrature points. The component evaluation reduces to tensor contractions:
\begin{equation}
T_j(z_j ; z_{<j}) \approx f_j(z_{<j}) + z_j \mathbf{w}^{\top} r(g_j(\mathbf{t} z_j, z_{<j}))
\end{equation}
This allows the integral for every site in a large batch to be estimated with a single, highly parallelized forward pass through the networks $f_j$ and $g_j$.
\paragraph{Why Neural Networks over Polynomials?} Neural networks provide a non-parametric, adaptive parameterization for the high-dimensional and a priori unknown target distributions of lattice theories. In contrast to fixed-degree polynomial expansions, they learn the required functional basis and complex conditional dependencies directly from the data. Universal approximation theorems formally guarantee that their expressive capacity is at least equivalent to that of polynomial maps.

\section{From Conditional Independence to Sparse Triangular Maps}
Our goal is to construct an efficient autoregressive (triangular) transport for a local target $P[\phi]$. Fix an ordering of lattice sites; by the chain rule,
\[
P[\phi] \;=\; \prod_{j=1}^{N} P(\phi_j \mid \phi_{<j}),\qquad
\phi_{<j}=(\phi_1,\ldots,\phi_{j-1}).
\]
Although $P$ is a local Markov random field, the \emph{exact} Knothe-Rosenblatt (KR) conditionals $P(\phi_j \mid \phi_{<j})$ are typically dense: conditioning on $\phi_{<j}$ implicitly marginalizes $\phi_{>j}$ and can induce long-range couplings. This densification is often illustrated by \emph{fill-in} in sparse eliminations (e.g., Cholesky for Gaussians) \citep{schäfer2021sparsecholeskyfactorizationkullbackleibler}, but the notions are distinct: KR conditionals are properties of the target density $P$ itself, whereas fill-in is a property of a chosen elimination scheme in a graphical/linear-algebraic factorization. Given a fixed coordinate ordering, the KR rearrangement uniquely specifies a lower-triangular monotone transport.

Finding an ordering that minimizes this fill-in is an NP-complete problem, rendering the construction of an exactly sparse map computationally intractable. To ensure scalability, we instead enforce approximate sparsity. This is achieved by restricting the dependencies of the $j$-th component of our triangular map, $\phi_j = T_j(z_j; \{\phi_i\}_{i \in \mathcal{C}(j)})$, to a local conditioning set $\mathcal{C}(j)$. Motivated by the physical locality of the action, we define this set as the "past neighbors" of site $j$:
$$\mathcal{C}(j) = N_p(j) \equiv N(j) \cap \{0, \dots, j-1\}$$
where $N(j)$ is the set of immediate neighbors of site $j$ on the lattice (for a concrete structure of the distribution, see \ref{markov_property}). This formulation presents a clear trade-off. By defining the conditioning context via $N_p(j)$, the size of the set is bounded by the lattice coordination number (e.g., $2D$), guaranteeing that the map evaluation scales linearly with the system volume, $\mathcal{O}(N)$. However, by ignoring the fill-in from marginalization, the map only approximates the true conditional structure. Consequently, to sample exactly from the target distribution, this approximate map must be used as a proposal within a Metropolis-Hastings correction framework. The quality of this approximation, and thus the overall sampling efficiency, critically depends on the chosen variable ordering.
\subsection{Impact of Variable Orderings on Sparsity Patterns}
\begin{figure}[tbp]
    \centering
    \includegraphics[width=\linewidth]{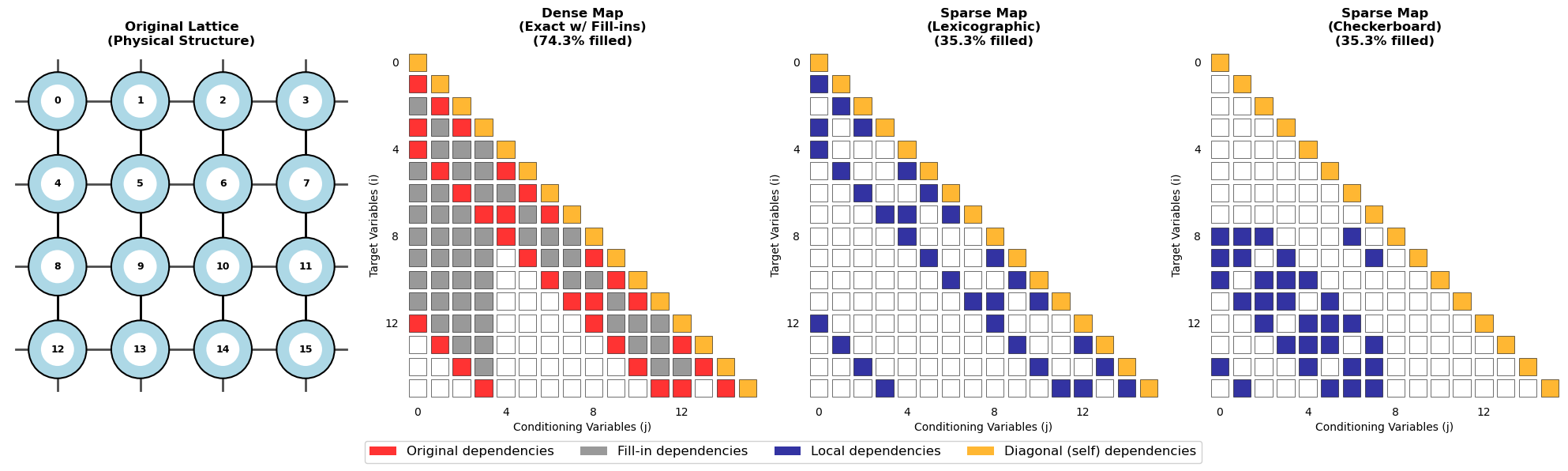}
    \caption{Dependency structures for a triangular map on a $4 \times 4$ lattice. It contrasts the dense, exact map for a lexicographic ordering (including "fill-in") with the enforced sparse maps for lexicographic and checkerboard ordering.}
    \label{fig:sparsity_structure}
\end{figure}
 The ordering determines which neighbors are "past" (and thus available as conditioning variables) and which are "future" (and must be marginalized over implicitly). We want an ordering that maximizes the information captured by the past neighbors. Our key insight is that while exact sparsity requires considering marginalization graphs with inevitable fill-ins during triangular decomposition, approximate sparsity based on conditional independence patterns leads to computationally efficient maps at the cost of reduced expressivity. This mirrors classical fill-in behavior in sparse Cholesky factorizations \citep{schäfer2021sparsecholeskyfactorizationkullbackleibler}. We investigate three strategies:
\paragraph{Lexicographic Ordering} Sites are ordered row-by-row: $\pi(i) = (i \mod L, \lfloor i/L \rfloor)$ for 2D lattices. The advantages are the simple implementation and predictable structure; however, it creates asymmetric dependencies, which are unfavorable for periodic boundaries.
\paragraph{Checkerboard Ordering} This strategy divides the lattice into two disjoint sets based on the parity of the sum of a site's coordinates. A site $x$ with coordinates $\left(c_1, \ldots, c_D\right)$ is even if $\sum_k c_k$ is even and odd otherwise. The ordering sequence is constructed in two stages: First, all even sites are ordered (typically lexicographically among themselves).Second, all odd sites are ordered. A key feature of this ordering is that all nearest neighbors of an even site are odd, and vise-versa. This means that when an odd site $\pi_j$ is processed, all of its nearest neighbors are guaranteed to be "past" neighbors, having already been processed in the first stage. This maximizes the available local information for the second half of the variables, making it highly effective for models based on nearest-neighbor interactions.
\paragraph{Max-Min Distance Ordering} This is a greedy, non-local method that constructs an ordered sequence of sites, $\pi=\left(\pi_1, \pi_2, \ldots, \pi_N\right)$, to break the spatial correlations between consecutive sites in the sequence. It aims to make each new site as physically far as possible from all previously chosen sites. Starting with an initial site (e.g., $\pi_1=0$ ), the subsequent sites are chosen iteratively. At step $j$, the algorithm selects the site $\pi_j$ from the set of remaining sites, $\Lambda \backslash\left\{\pi_1, \ldots, \pi_{j-1}\right\}$, that maximizes the minimum distance to any already-ordered site:
$$
\pi_j=\arg \max _{x \in \Lambda \backslash\left\{\pi_1, \ldots, \pi_{j-1}\right\}}\left(\min _{k<j} d\left(x, \pi_k\right)^2\right)
$$
Here, $d(x, y)^2$ is the squared Euclidean distance between sites $x$ and $y$, calculated with periodic boundary conditions. This approach creates a more uniform distribution of "past" neighbors across the lattice, avoiding the directional bias of the lexicographical ordering.

Periodic Boundary Conditions (PBCs) introduce topological complexity. Sites that are physically adjacent may be far apart in the ordering due to boundary wrap-around (e.g., site 0 and site $L-1$ in 1D). We rely on the flexibility of the MRNN parameterization and the subsequent MCMC correction to ensure asymptotical exactness. The effect of the different orderings on the structure of the triangular map for a lattice problem can be seen in Figure \ref{fig:sparsity_structure}. Here, we represented a $L=4$ lattice with periodic boundary conditions and show the difference between the exact conditional independence and the enforced sparse representation. A more detailed analysis of the scaling behavior of fill-ins and an overall overview of the effect of the different orderings can be seen in \ref{fill_in_scaling}.

\section{Experiments}
We evaluate our proposed sparse triangular maps on the 2D $\phi^4$ theory ($D=2$). The map parameters are optimized by minimizing the variational free energy, $\mathcal{L}(\theta) = \mathbb{E}_{z \sim p_Z(z)} [S[T_{\theta}(z)] - \log |\det J_{T_{\theta}}(z)|]$, via stochastic gradient descent (see \ref{kl_training}). This is equivalent to minimizing the reverse Kullback-Leibler divergence between the model distribution $p_{\Phi}$ and the target Gibbs distribution $P[\phi]$. To ensure exact sampling from the target distribution, we use the trained map as a proposal within an Independent Metropolis-Hastings (IMH) algorithm \citep{albergo2019flow} (see \ref{mh_step}).

\textbf{Setup.} We consider an $L=8$ lattice ($N=64$ sites) with periodic boundary conditions. The $\phi^4$ theory parameters are fixed at $m_0^2 = -4.0$ and $\lambda_0 = 8.0$. In this initial work, the parameters were chosen to lie in the symmetric phase. The primary performance metric is the effective sample size (ESS) (see \ref{app:ess_definition}).

\textbf{Model and Training.} Unless specified otherwise, maps are constructed from Monotone Rectified Neural Network (MRNN) components. The neural network for each component $T_j$ consists of 3 hidden layers (64 units each, GELU activation). Monotonicity is enforced via a Softplus activation function in the final layer of the integrand network, and the required integral is approximated using a 15-point Gauss-Legendre quadrature. All models were trained for 3000 epochs using the AdamW optimizer (initial LR $10^{-3}$, weight decay $10^{-5}$) with a batch size of 256. A cosine annealing schedule decayed the learning rate to a minimum of $10^{-6}$.

\subsection{Impact of Variable Ordering and Neighborhood Size}

We investigate the fundamental trade-off between the sparsity of the triangular map and its expressivity by analyzing how different variable ordering strategies and conditioning neighborhood sizes affect performance.

We evaluated three distinct ordering strategies: a standard \textbf{Lexicographical} ordering, a physics-motivated \textbf{Checkerboard} (CB) ordering, and a \textbf{MaxMin} ordering designed to maximize spatial separation between causally dependent variables. For each ordering, we systematically increase the map's complexity by cumulatively expanding the conditioning neighborhood: \textbf{1st-order} (nearest-neighbors), \textbf{2nd-order} (including diagonals), and \textbf{3rd-order} (including "knight-moves").

\begin{figure}[tbp]
    \centering
    \includegraphics[width=\linewidth]{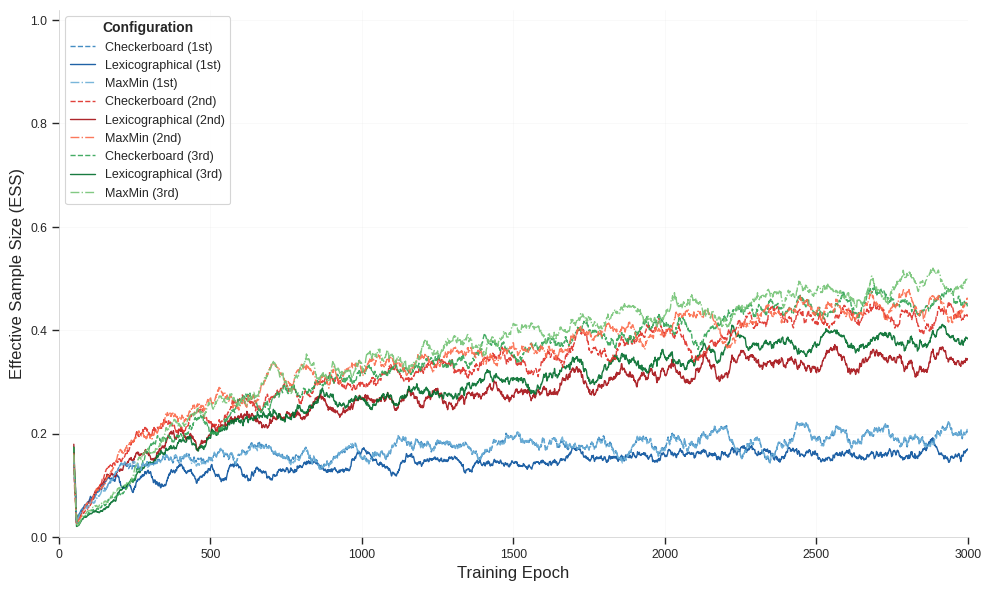}
    \caption{Effective Sample Size (ESS) as a function of training epochs for the nine model configurations. Each panel corresponds to a different variable ordering strategy (Lexicographical, Checkerboard, MaxMin). Within each panel, lines represent models trained with cumulatively increasing neighborhood orders (1st, 2nd, and 3rd). Expanding the conditioning set consistently improves sampling efficiency, with the MaxMin ordering achieving the highest overall performance.}
    \label{fig:map_sparsity}
\end{figure}

We analyzed these nine configurations based on the realized sparsity (Avg.~$|C(j)|$) and sampling efficiency (ESS). As visualized in Figure~\ref{fig:map_sparsity}, increasing the neighborhood order consistently improves the final ESS across all orderings, demonstrating that a richer local context allows the model to better capture the underlying physics. The MaxMin ordering achieves the highest overall performance (although it is similar in performance to the Checkerboard ordering), confirming that its structure naturally better preserves long-range interactions, providing a more effective conditioning context compared to the other strategies.

\subsection{Architecture Comparison and Scalability}

We perform a direct benchmark of various flow-based architectures to determine the most effective and computationally efficient design for lattice field theory sampling. We compare our triangular transport models against a convolutional RealNVP baseline.

Five architectures are compared: (1) A \textbf{Dense} triangular map (lexicographical ordering), serving as an upper-bound for expressivity while ignoring conditional independence. (2) A \textbf{RealNVP (CNN)} model, built from 8 coupling layers, is representative of standard flow models for structured data. (3-5) \textbf{Sparse Triangular Maps} use MaxMin ordering, with dependencies restricted to 1st-order neighbors, 2nd-order neighbors, and the \textbf{Exact Conditional} dependencies derived from graph elimination.

\begin{figure}[htbp]
    \centering
    \begin{subfigure}[b]{0.51\textwidth}
        \centering
        \includegraphics[width=\linewidth]{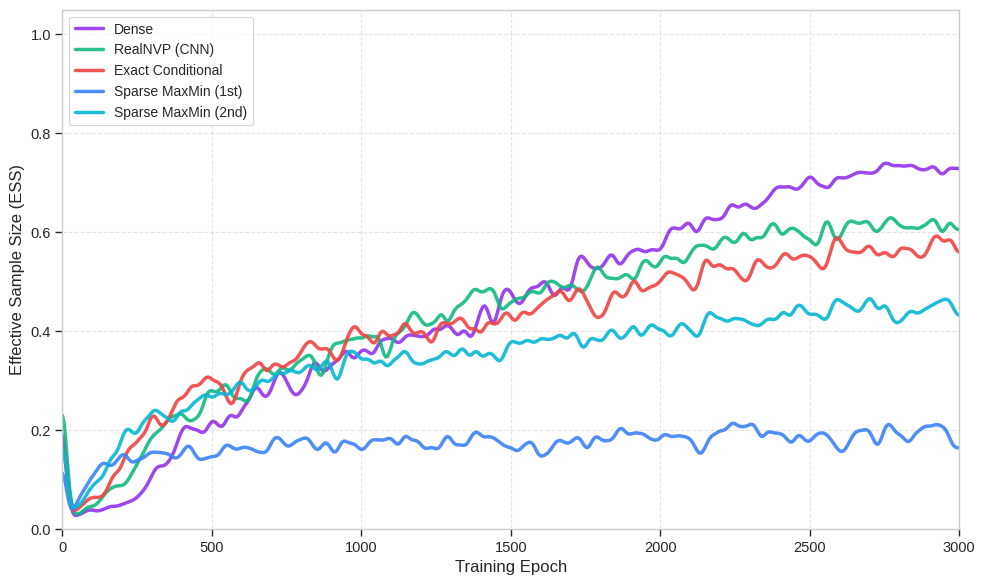}
        \caption{ESS evolution during training.}
        \label{fig:ess_evolution}
    \end{subfigure}
    \hfill
    \begin{subfigure}[b]{0.47\textwidth}
        \centering
        \includegraphics[width=\linewidth]{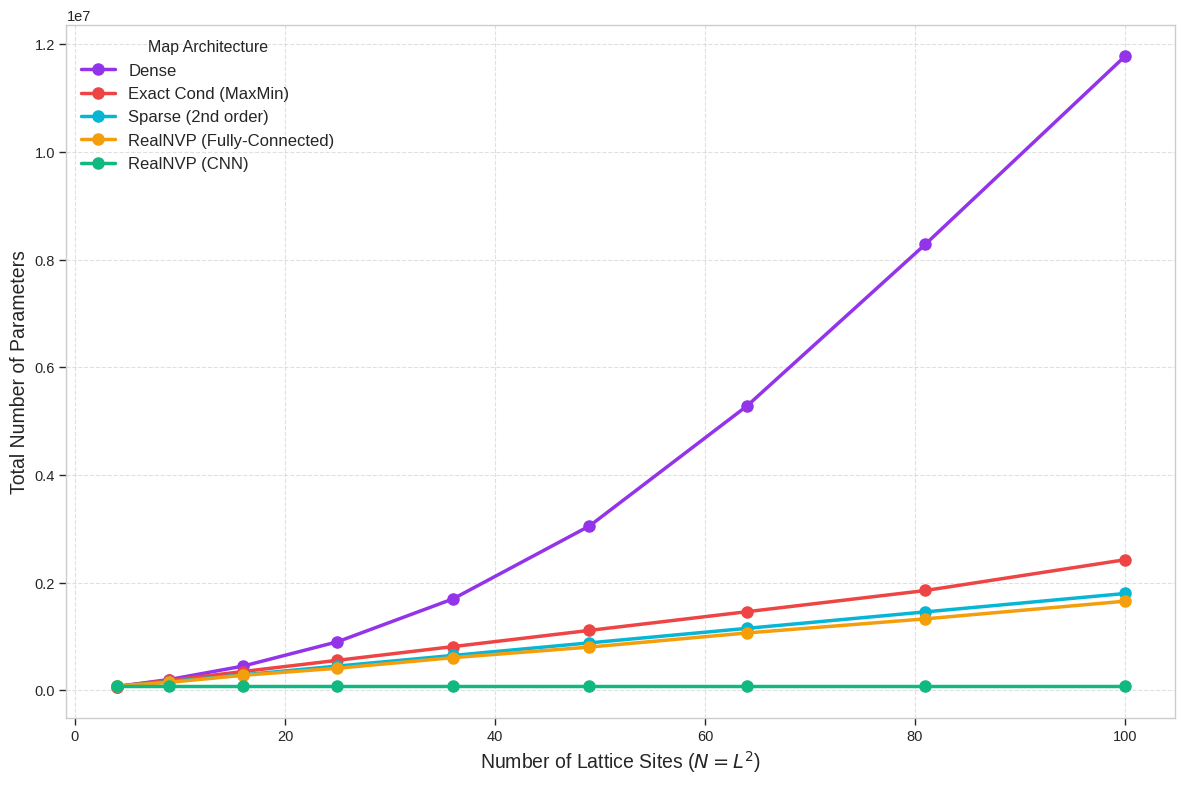}
        \caption{Parameter Scaling}
        \label{fig:final_ess}
    \end{subfigure}
    \caption{Performance comparison of various flow-based architectures. Subfigure (a) plots the ESS during training. Subfigure (b) summarizes the parameter scaling based on the number of lattice sites}
    \label{fig:flow_comparison}
\end{figure}

The results are summarized in Figure~\ref{fig:flow_comparison}. The CNN-based RealNVP performs competitively, achieving an ESS comparable to that of the exact conditional map. The approximate sparse maps (2nd order) perform slightly worse, highlighting the challenge of hand-picking an optimal, fixed-size conditioning set.

Convolutional RealNVP is parameter-efficient; its trainable weights do not grow with the lattice size $N$. Inference, however, processes the entire configuration through $L$ coupling layers. While each layer exploits spatial parallelism across sites and channels, the $L$ layers must be evaluated in sequence, making performance depth-limited and causing compute and activations to scale linearly with $N$.

In contrast, while the sparse MRNN's parameter count grows linearly $\mathcal{O}(N)$, its architecture is parallelizable across sites. Each of the $N$ map components is an independent neural network that requires only the base sample $z_j$ and its small, local conditioning set $\{z_k\}_{k \in C(j)}$. This structure allows for spatial parallelism, where all $N$ components can be computed simultaneously. Therefore, the MRNN's improved parallelization potential makes it a valid alternative for large lattices.

\subsection{Statistical Error and Physical Observables}

We analyze the statistical error of physical observables as a function of the number of generated samples, comparing our optimized triangular map against a standard Hybrid Monte Carlo (HMC) sampler.

The HMC sampler uses a standard leapfrog integrator with 10 steps. The integrator step size, $\epsilon$, was dynamically tuned to achieve a target acceptance rate of $\approx 70\%$. The implementation correctly handles the lattice's periodic boundary conditions by using circular shifts in the force term computation. The triangular map, used within the IMH framework, was tuned for proposal scale to achieve $\sim 50 \%$ acceptance, following common practice in flow-corrected IMH \citep{albergo2019flow}).

\begin{figure}[htbp]
    \centering
    \includegraphics[width=\linewidth]{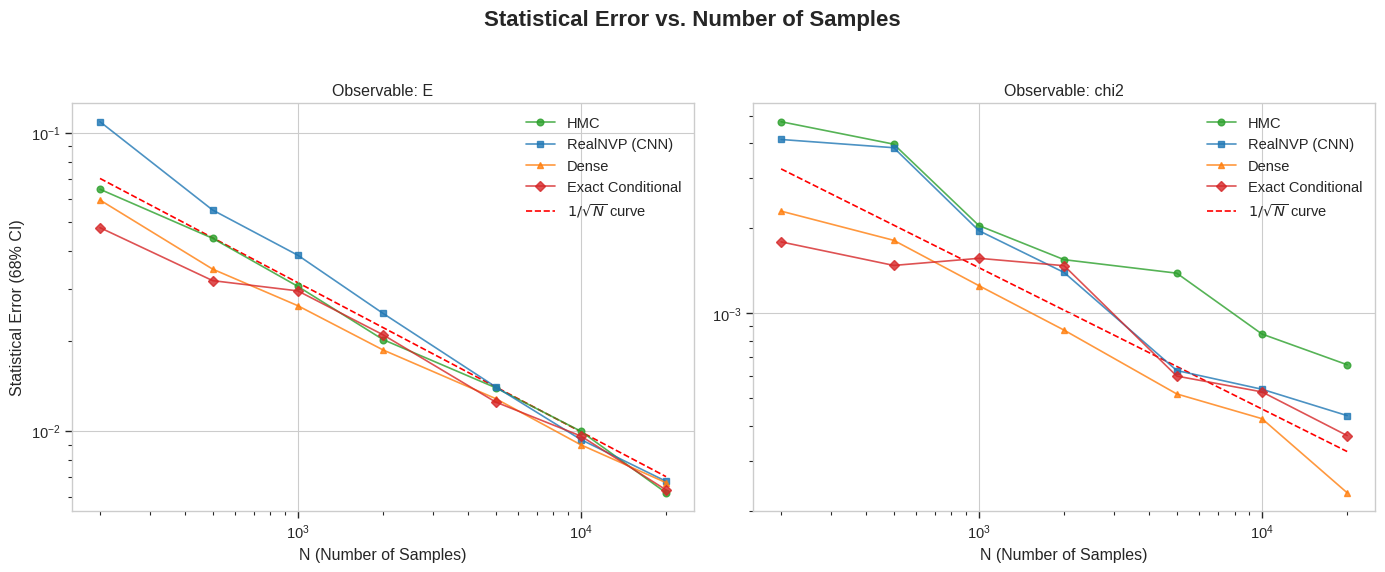}
    \caption{The statistical error of the measured energy and susceptibility as a function of the number of samples, $N$. The plot compares HMC to the transport maps. The solid lines show the statistical error, while the red dashed line represents the theoretical $1/\sqrt{N}$ behavior of an ideal sampler.}
    \label{fig:statistical_error}
\end{figure}

For each sampler, we generated a chain of 20,000 configurations, discarding the first 2000 samples as burn-in. We measured the energy $\langle E \rangle$ and the susceptibility $\chi_2$ (for the exact definition, see \ref{app:observables}). The statistical error was estimated using the bootstrap method (500 resamples, 68\% confidence interval) for varying sub-sample sizes ($M=200$ up to $M=20,000$). The results presented in Figure~\ref{fig:statistical_error} confirm that the triangular map closely follows the ideal $1/\sqrt{N}$ scaling behavior, achieving a lower statistical error for a given number of samples than the HMC method  for susceptibility.

\section{Conclusion and Future Outlook}
This work introduces a highly scalable framework for sampling in lattice field theories using triangular transport maps. By leveraging the inherent locality of the physical action, we construct maps with $\mathcal{O}(N)$ complexity, a significant improvement over the $\mathcal{O}(N^2)$ scaling of dense autoregressive models. We propose a principled framework to manage the trade-off between precision and computational performance. We demonstrate that, while an exact triangular map requires modeling dense, non-local dependencies ("fill-in") created by marginalization, an approximate sparse map—conditioned only on local, preceding physical neighbors—provides a powerful and scalable alternative.

Any error introduced by the sparsity approximation is corrected by a final Metropolis-Hastings step, guaranteeing that the resulting samples are drawn from the exact Boltzmann distribution. Our experiments on the 2D $\phi^4$ theory show promising initial results. We systematically show that physics-informed orderings, such as the checkerboard or MaxMin pattern, outperform simpler ones. The method produces physical observables with lower statistical error for a given number of samples, closely tracking the ideal $1/\sqrt{N}$ scaling and confirming its practical advantage, especially the parallelizability, for physics simulations.

Our framework establishes an alternative foundation for parallelizable samplers in lattice field theories, with several exciting avenues for future work.

\paragraph{Limitations} Triangular transport maps require the latent space dimension to match the configuration space dimension exactly, preventing dimensionality reduction that could accelerate sampling in systems with redundant degrees of freedom. The method has not been tested near critical points where correlation lengths diverge. Importantly, our experiments are restricted to small lattices in the symmetric phase, limiting the generalizability of our performance claims to larger, physically relevant scales.  Further investigation is required to validate these findings on larger lattices and in the critical regimes.

\paragraph{Future Outlook} The most significant frontier is the application to non-Abelian gauge theories. In these theories, variables are elements of a Lie group (e.g., SU(N)) associated with the links of the lattice, and the action is constructed from gauge-invariant objects. For example, the Wilson gauge action is built from plaquette variables $U_\square$:
\begin{equation}
S[U] = \beta \sum_{\square} \text{Re}\,\text{Tr}(1 - U_\square)
\end{equation}
Extending our framework requires developing gauge-equivariant triangular maps on the SU(N) group manifold that rigorously preserve local gauge symmetry. A promising path involves designing gauge-equivariant MRNNs  that condition on local, gauge-covariant stencils (e.g., small Wilson loops). This would involve parameterizing transformations in the Lie algebra via exponential coordinates and incorporating corrections for the Haar measure to ensure the map is properly defined on the group.

While we have shown the power of sparsity, there is room for further optimization. Instead of relying on fixed, predefined orderings, it may be possible to learn an ordering; however, though naively this adds a combinatorial layer that must be amortized or relaxed.

\paragraph{Extensions to fermionic field theories.}
A natural extension arises in lattice QCD, where configurations comprise bosonic gauge links $\phi\in \mathrm{SU}(N)$ and fermionic Grassmann fields. Integrating out the Grassmann variables yields
\[
P[\phi]\ \propto\ e^{-S_{\text{gauge}}[\phi]}\,\det D[\phi],
\]
where $S_{\text{gauge}}[\phi]$ is the gauge action (e.g., Wilson) and $D[\phi]$ is the Dirac operator. Introducing complex pseudofermion fields $\chi$ gives the standard representation
\[
\det\!\big(D^\dagger D\big)[\phi]\ \propto\ \int \mathcal{D}\chi\;\exp\!\Big(-\,\chi^\dagger (D^\dagger D)^{-1}[\phi]\,\chi\Big),
\]
so the joint measure factorizes as
\[
P[\phi,\chi]\;=\;P_{\text{gauge}}[\phi]\;P_{\text{fer}}(\chi\,|\,\phi),\qquad
P_{\text{gauge}}[\phi]\propto e^{-S_{\text{gauge}}[\phi]},\quad
P_{\text{fer}}(\chi\,|\,\phi)=\mathcal{N}_{\mathbb{C}}\!\big(0,\;C[\phi]\big),
\]
with covariance $C[\phi]=(D^\dagger D)^{-1}[\phi]$ and precision $M[\phi]=D^\dagger D[\phi]$. Accordingly, a triangular transport decomposes block-triangularly:
\[
T^{-1}(\phi,\chi)\;=\;
\begin{bmatrix}
T^{-1}_{\text{gauge}}(\phi)\\[0.2em]
T^{-1}_{\text{fer}}(\chi;\,\phi)
\end{bmatrix}
=
\begin{bmatrix}
z_{\text{gauge}}\\[0.2em]
z_{\text{fer}}
\end{bmatrix},
\qquad
\text{so sampling uses }\ \phi=T_{\text{gauge}}(z_{\text{gauge}}),\ \chi=T_{\text{fer}}(z_{\text{fer}};\phi),
\]
which preserves the block-triangular Jacobian
$\bigl[\begin{smallmatrix} T_{\text{gauge}} & 0 \\ * & T_{\text{fer}} \end{smallmatrix}\bigr]$
and enables conditional triangular maps for pseudofermions. For gauge fields, maps must be defined on $\mathrm{SU}(N)$ with respect to the Haar measure (e.g., via Lie-algebra charts with the appropriate Jacobian). Alternative approaches to incorporating fermions in flow models have also been explored in \citep{Albergo_2021_fermions}.
%%%%%%%%%%%%%%%%%%%%%%%%%%%%%%%%%%%%%%%%%%%%%%%%%%%%%%%%%%%%
\paragraph{Reproducibility.} The source code is available at \url{https://github.com/andreyb18/transport_maps_for_lattice_qcd.git}.

\section*{Acknowledgments}
AB and YM are supported in part by the US Department of Energy, SciDAC-5 program, under contract DE-SC0012704, subcontract 425236. We thank Daniel Sharp for very insightful and helpful discussions.

\bibliography{references}
\bibliographystyle{plain}
\appendix

\section{Supplementary Material}
\subsection{Training: KL Divergence Minimization}
\label{kl_training}
The parameters of the neural networks modeling $f_j$ and $g_j$ (collectively denoted $\theta$) are trained by minimizing the Kullback-Leibler (KL) divergence between the model distribution $p_{\Phi}(\phi)$ (induced by $T_\theta$) and the target distribution $P[\phi]$ (Eq. \eqref{eq:boltzmann}). We minimize $KL(p_{\Phi}(\phi) || P[\phi])$:
\begin{align}
KL(p_{\Phi} || P) &= \int p_{\Phi}(\phi) \log \frac{p_{\Phi}(\phi)}{P[\phi]} d\phi \\
&= \mathbb{E}_{\phi \sim p_{\Phi}} [\log p_{\Phi}(\phi) - \log P[\phi]]
\end{align}
This is the appropriate loss (reverse KL or variational free energy minimization) when the target density is known (via the action $S[\phi]$) but samples are unavailable. Using the change of variables $\phi = T_{\theta}(z)$ where $z \sim p_Z(z)$ (the base Gaussian distribution), and $p_{\Phi}(T_{\theta}(z)) = p_Z(z) |\det J_{T_{\theta}}(z)|^{-1}$:
\begin{align}
KL(p_{\Phi} || P) &= \mathbb{E}_{z \sim p_Z(z)} [\log (p_Z(z) |\det J_{T_{\theta}}(z)|^{-1}) - \log (Z^{-1} e^{-S[T_{\theta}(z)]})] \\
&= \mathbb{E}_{z \sim p_Z(z)} [\log p_Z(z) - \log |\det J_{T_{\theta}}(z)| + S[T_{\theta}(z)] + \log Z]
\end{align}
To minimize this KL divergence, we can drop terms constant with respect to model parameters $\theta$ (namely $\mathbb{E}_{z \sim p_Z(z)}[\log p_Z(z)]$ and $\log Z$). The loss function to minimize is thus:
\begin{equation}
\mathcal{L}(\theta) = \mathbb{E}_{z \sim p_Z(z)} [S[T_{\theta}(z)] - \log |\det J_{T_{\theta}}(z)|]
\label{eq:loss_function}
\end{equation}
which in this setup becomes:
\begin{equation}
\mathcal{L}(\theta) = \mathbb{E}_{z \sim p_Z(z)} \left[ S[T_{\theta}(z)] - \sum_{j=0}^{N-1} \log r(g_j(z_j, z_{<j}^{(j)}; \theta)) \right]
\end{equation}
where $z_{<j}^{(j)}$ denotes the appropriate conditioning set for the $j^{th}$ component (all $z_{<j}$ for dense, or $z_i$ for $i \in N_p(j)$ for sparse). The expectation is approximated by Monte Carlo sampling from $p_Z(z)$ and using mini-batch stochastic gradient descent. 

\subsection{Metropolis-Hastings Correction Step}
\label{mh_step}
While the trained normalizing flow $p_{\Phi}(\phi)$ approximates $P[\phi]$, it may not be exact. To obtain samples from the exact target distribution $P[\phi]$, an MCMC correction step is applied. Similar to \citep{albergo2019flow}. We use the Independent Metropolis-Hastings (IMH) algorithm, where the proposal distribution is the learned map itself, $Q(\phi') = p_{\Phi}(\phi')$. Given a current sample $\phi_c$, a new sample $\phi_p$ is proposed by drawing $z_p \sim p_Z(z)$ and setting $\phi_p = T_{\theta}(z_p)$. The acceptance probability is:
\begin{equation}
\alpha(\phi_p | \phi_c) = \min \left(1, \frac{P[\phi_p] Q(\phi_c)}{P[\phi_c] Q(\phi_p)} \right) = \min \left(1, \frac{P[\phi_p] p_{\Phi}(\phi_c)}{P[\phi_c] p_{\Phi}(\phi_p)} \right)
\end{equation}
This can be rewritten using importance weights $w(\phi) = P[\phi]/p_{\Phi}(\phi)$: $\alpha(\phi_p | \phi_c) = \min \left(1, \frac{w(\phi_p)}{w(\phi_c)} \right)$. The log-importance weight for a sample $\phi = T_{\theta}(z)$ is (ignoring the constant $\log Z$):
\begin{align}
\log w'(\phi) &= \log (e^{-S[\phi]}) - \log p_{\Phi}(\phi) \\
&= -S[T_{\theta}(z)] - (\log p_Z(z) - \log |\det J_{T_{\theta}}(z)|)
\end{align}
The ratio $w(\phi_p)/w(\phi_c)$ becomes $w'(\phi_p)/w'(\phi_c)$, and the acceptance probability calculation proceeds using these relative weights. The MCMC step ensures that the resulting chain of accepted samples converges to the exact target distribution $P[\phi]$.

\subsection{Effective Sample Size (ESS) Definition}
\label{app:ess_definition}
The quality of the approximation $p_{\Phi}$ can be measured by the Effective Sample Size (ESS) of the samples generated directly from the flow, using importance weights $w(\phi) = P[\phi]/p_{\Phi}(\phi)$. For $M$ samples $\{\phi_i\}_{i=1}^M$:
\begin{equation}
\text{ESS} = \frac{(\sum_{i=1}^M w(\phi_i))^2}{\sum_{i=1}^M w(\phi_i)^2} / M
\end{equation}
An ESS close to 1 indicates that $p_{\Phi} \approx P$.
\subsection{Markov Property of $\phi^4$ theory}
\label{markov_property}
The Markov property emerges from the locality of the action. For the $\phi^4$ theory, the full conditional is:

$$
\begin{aligned}
P\left(\phi_x \mid \phi_{\Lambda \backslash\{x\}}\right) & =\frac{P[\phi]}{P\left[\phi_{\Lambda \backslash\{x\}}\right]} \\
& \propto \exp \left(-\sum_\mu \frac{1}{2}\left(\phi_{x+\hat{\mu}}-\phi_x\right)^2-\frac{m_0^2}{2} \phi_x^2-\frac{\lambda_0}{4!} \phi_x^4\right)
\end{aligned}
$$

This depends only on $\phi_{x \pm \hat{\mu}}$, confirming $\phi_x \perp \phi_{\Lambda \backslash(\{x\} \cup \mathcal{N}(x))} \mid \phi_{\mathcal{N}(x)}$.

\subsection{Triangular Map Structures and Fill-in Scaling}
\label{fill_in_scaling}
\begin{figure}[htbp]
    \centering
    \includegraphics[width=\linewidth]{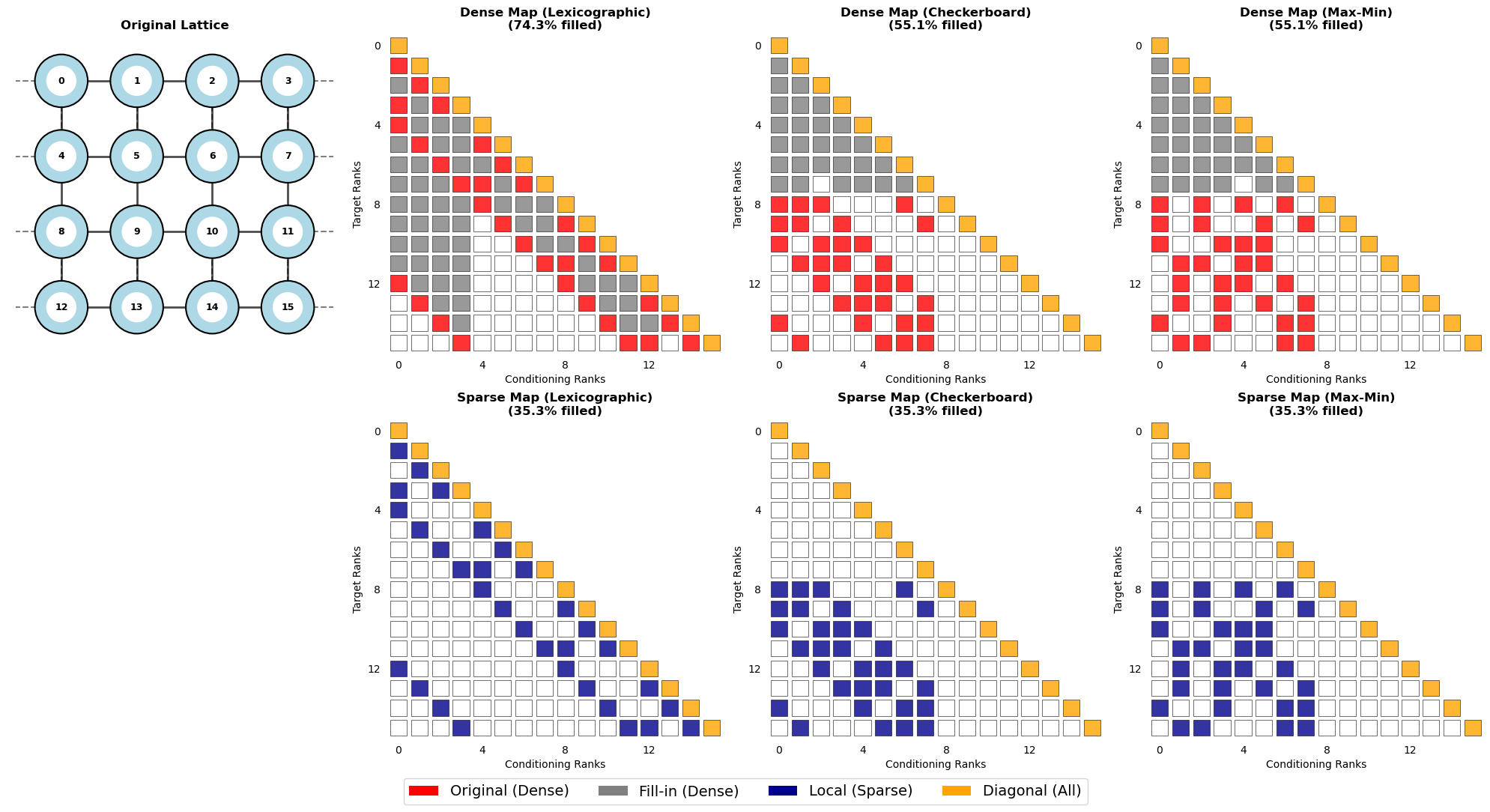}
    \caption{Comparison of exact (top row) and enforced sparse (bottom row) dependency structures for a triangular map on a $4 \times 4$ lattice under different orderings. The exact maps reveal the non-local fill-in patterns unique to each ordering, with lexicographic ordering creating a distinctly different dense structure from the more symmetric checkerboard ordering. The sparse maps are, by construction, limited to preceding physical neighbors ($N_p(j)$), highlighting the significant reduction in complexity at the cost of approximation.}
    \label{fig:appendix_sparsity_comparison}
\end{figure}

\begin{figure}[htbp]
    \centering
    \includegraphics[width=0.5\linewidth]{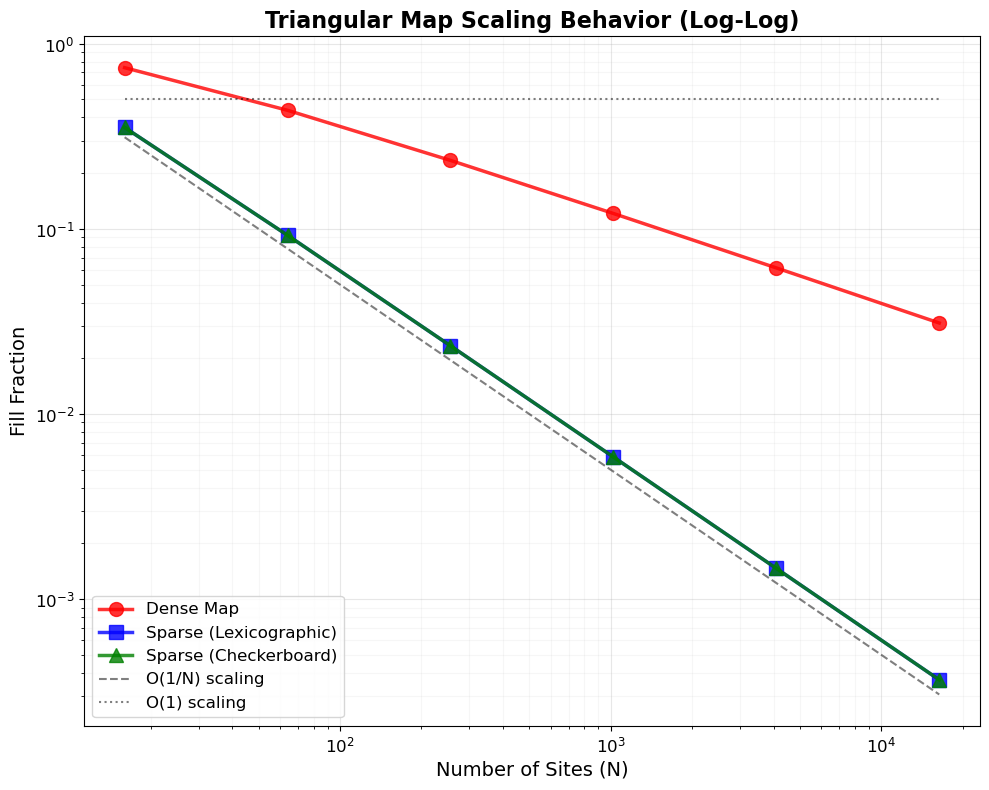}
    \caption{This plot illustrates the computational challenge of using an exact conditional map (Dense Map) on larger 2D lattices. We measure its "fill-in" rate, the growth of non-local dependencies, which exceeds the ideal $\mathcal{O}(1 / N)$ scaling. This motivates the use of computationally efficient Sparse (Lexicographic) and Sparse (Checkerboard) approximations.
    }
    \label{fig:appendix_fill_in_scaling}
\end{figure}

The ordering of variables is a critical choice in constructing triangular maps, as it directly dictates the structure of both the exact and approximate dependency graphs. The exact dependency structure, required for a perfect transformation, accounts for all correlations induced by marginalizing "future" variables. This process, known as fill-in, typically results in a dense graph. In contrast, an approximate map achieves computational efficiency by enforcing a sparse structure based only on local, "past" physical neighbors. In Figure \ref{fig:appendix_sparsity_comparison} and Figure \ref{fig:appendix_fill_in_scaling} , we provide a detailed visualization of these effects and analyze the scaling behavior of the fill-in phenomenon for different orderings.
\section{Validation of Physical Observables}
\label{app:observables}

To validate the asymptotical exactness of the sampling procedure (flow + IMH), we will calculate key physical observables and compare them against HMC results.
\begin{itemize}
    \item \textbf{Average Magnetization:} $\langle M \rangle = \langle |\frac{1}{N}\sum_x \phi_x| \rangle$.
    \item \textbf{Magnetic Susceptibility:} $\chi_2 = N (\langle M^2 \rangle - \langle M \rangle^2)$.
\end{itemize}
%%%%%%%%%%%%%%%%%%%%%%%%%%%%%%%%%%%%%%%%%%%%%%%%%%%%%%%%%%%
\end{document}